# Comparison of Several Sparse Recovery Methods for Low Rank Matrices with Random Samples

Ashkan Esmaeili
ACRI and EE Dept.
Sharif University of Technology
Tehran, Iran
aesmaili@stanford.edu

Farokh Marvasti
ACRI and EE Dept.
Sharif University of Technology
Tehran, Iran
marvasti@sharif.edu

*Abstract*— In this paper, we will investigate the efficacy of IMAT (Iterative Method of Adaptive Thresholding) in recovering the sparse signal (parameters) for linear models with missing data. Sparse recovery rises in compressed sensing and machine learning problems and has various applications necessitating viable reconstruction methods specifically when we work with big data. This paper will focus on comparing the power of IMAT in reconstruction of the desired sparse signal with LASSO. Aditionally, we will assume the model has random missing information. Missing data has been recently of interest in big data and machine learning problems since they appear in many cases including but not limited to medical imaging datasets, hospital datasets, and massive MIMO. The dominance of IMAT over the well-known LASSO will be taken into account in different scenarios. Simulations and numerical results are also provided to verify the arguments.

## I. INTRODUCTION

In this paper, we investigate the efficacy of different methods in recovering the sparse parameters signal for the scenario when the dataset contains missing entries. We want to find out which method is more powerful in recovering the parameters while there are missing samples. We consider the Lasso method which is well-known for sparse recovery as well as IMAT, and IHT. IMATCS is a modified version of IMAT [1] which will be used throughout the paper as a method for retrieving the compressed sensing solution. IHT(Iterative Hard Thresholding) is another sparse recovery method which has access to the degree of sparsity in the parameters signal. There has been lots of work done in the CS, Signal Processing, and Machine Learning literature highlighting the aforementioned problem. The most well-known approach known to statisticians facing the above problem is to apply $l_1$ norm penalty term to the least square objective function and solving that using efficient algorithms in literature. We will consider two different cases in this paper. First, we simply compare the reconstruction quality of these methods without any matrix completion on missing data. Then, we also consider them for the case where we initially apply the matrix completion, and then we compare the efficacy of the three methods. There are plenty of matrix completion methods introduced in the literature such as Optspace [5], SVD Regression, and Soft Thresholding [4]. We will focus on one of the methods called Soft Thresholding whose complexity is in computing the SVD in each iteration. This is for low rank data reconstruction but the results of paper are general, and we are considering both low rank and high rank data in our simulations section. However; in the section which we precomplete the data initially, we only foucs on low rank matrix completion. Low rank models are of interest because they have numerous applications. It is worth mentioning that we have modified the approach of choosing the thresholds in comparison to the IMATCS in [1]. Actually, we are using an adaptive thresholding method depending on the energy of signal in the previous iteration. The intuition behind choosing the thresholds will be elaborated upon later on in the paper. We will also include the IHT (Iterative hard thresholding) method and show the superiority of IMAT in comparison to IHT.

## II. PROBLEM MODEL

We consider the problem of finding the sparse signal $\beta$ in the following true linear model:

$$Y = X\beta + \epsilon \qquad (1)$$

where, $X \in R^{m \times n}$ is the data matrix, $\beta \in R^n$ is the parameters signal, $\epsilon \sim N(0, I_{n \times n})$ is the i.i.d noise, and $Y \in R^m$ is the observed label vector.

We assume $\beta$ is sparse meaning that the nonzero number of elements in $\beta$ is $s \ll n$. The support of $\beta$ is defined as follows:

$$Supp(\beta) = \{i \in \{1, \dots, n\} : \beta(i) \neq 0\} \qquad (2)$$

Therefore $|Supp(\beta)| = s$, where $s \ll n$. We also suppose that $X$ has missing entries. For example, we can assume $X$ is generated from an oracle $\tilde{X}$ as follows:

$$X = \tilde{X} \odot B, where\ B_{i,j} = Ber(\alpha) \qquad (3)$$

After introducing the problem model we will briefly explain the approaches taken into account for sparse recovery.

## A. Lasso

The logic of Lasso is applying $l_1$-norm penalty term as stated before to yield sparse solutions. Therefore, it is equivalent to finding the solution to the following minimization problem.

$$\beta^*(\lambda) = \min_{\beta} \left\|\hat{X}\beta - Y\right\|^2 + \lambda\|\beta\|_1 \qquad (4)$$

By cross-validating over $\lambda$'s grid and picking the optimal $\lambda$, the sparse signal is recovered.

## B. IMATCS

In this section, the proposed Iterative Method for Compressed Sensing recovery (IMATCS) is illustrated. IMATCS is an efficient method in finding the solution to the copmpressed sensing problem and it is a modified version of IMAT. It works iteratively as statetd in [1]. Briefly, we will explain how the IMATCS works. The mathematical formulation of the method is as follows:

$$\beta_{k+1} = T_{k+1}\left(\beta_k + \lambda X^H(Y - X\beta_k)\right) \qquad (5)$$

$$T_k = T_0 \exp(-\alpha k) \qquad (6)$$

Where the index $k$ denotes the number of the iteration. $\lambda$ is IMATCS parameter and is a determining factor for speed of convergemce. The equation (5) means that all values in the signal smaller than the threshold found in the last iteration are set to zero and the other values are kept. However in our version the adaptive thresholding taken into account is as follows :

$$T_{k+1} = c \times \overline{\beta_k} \qquad (7)$$

where $\overline{\beta_k}$ is the average of the signal $\beta_k$.

We are using a modified version of adaptive thresholding than what was already used in this method. The exponential thresholding which was decaying with the number of iterations is not used here. What we are using is proportional to the average of the signal used in the previous iteration. The intuition in using this thresholding is that the information of the distribution of the parameters signal exists in the retrieved signals of the previous iterations, and therefore we can learn about the structure of the desired signal throughout the iterations, intuitively based on Law of Large Numbers(LLN). Thus, we have the degree of freedom to tune the level of certainty in picking up the components we need in recovering the signal. The parameter $c$ is chosen by cross-validation on the training set. In fact, good choice of $c$ along with the information of the average signal recovered helps in picking the main components. Setting a small $c$ leads to adding noisy components and a non-sparse solution. Conversly, picking a large $c$ will lead to losing the main components and as a result strong bias which leads to error.

## C. IHT

We also provide the results of IHT (Iterative Hard Thresholding) in our simulations and compare the method results with those of IMAT and again show how IMAT is performing better for missing scenarios. The dynamics of IHT is similar to the IMAT to some extent except that the thresholding is done by selecting the $s$ largest components of signals retrieved at each iteration. This method is dependent on knowing the number of non-zero entries of the desired signal.

## III. MATRIX COMPLETION

In the second part of our analysis, we will first try to impute the missing entries in our data and from there on we will again apply the sparse recovery methods and compare their efficiency. Therefore, we first briefly mention some facts and points about the matrix completion here. In case $X$ has missing entries as described in the above, which happens in many scenarios like hospital patients data and massive MIMO datasets, we also need to recover the matrix with $X$ as $\hat{X}$. Thus, we have two phases of matrix completion followed by $l_1$ norm regularized least square minimization or IMAT.

Based on the missing structure of the matrix, many completion methods like Soft-Thresholding, Singular Value Thresholding, Optspace, and Nonconvex Factorization exist in the literature. These methods are usually complex and difficult in implementation. In this paper, we will consider low rank model for the matrix in the second part and then, we apply soft thresholding method to have an approximation of our data and from there on, we will apply the aforementioned sparse recovery methods and compare the results. Here, we briefly mention how soft thresholding method works. The completion problems for low rank cases usually solve the following minimization problem:

$$\min_{X} \left\|P_E(X - \hat{X})\right\|_2^2 + \lambda\|X\|_* \qquad (8)$$

where the second term is trace norm and the first term is residual on observed entries. The soft-thresholding algorithm for finding out the solution works iteratively as follows:

$$Z_{k+1} = Z_k \odot (1 - B) + X \odot B \qquad (9)$$
$$Z_{k+1} = USV^T \qquad (10)$$
$$Z_{k+1} = U(S - \lambda_1 I)V^T \qquad (11)$$

At each iteration an SVD is computed followed by thresholding. Since the main focus of this paper is on IMAT and not the completion methods, we refrain from including further details of the completion algorithm here.

Therefore, we have the following steps for the lasso method in this section: First,

$$X^*(\lambda) = argmin\left(\left\|P_E(X - \hat{X})\right\|_2^2 + \lambda\|X\|_*\right) \qquad (12)$$

In (12), the original data matrix which is assumed to be low-rank is recovered at the first stage by minimizing the residul defined as the norm-2 difference between the observed entries plus a penalty term which is minimizing the trace norm of our data assuming it is low rank. The minimizer of this problem after cross-validation over $\lambda$ values is assumed to be completed data. Now, we can apply lasso to the completed matrix as follows: Second,

$$\beta^*(\lambda_1) = \min_{\beta}\left\|\hat{X}\beta - Y\right\|^2 + \lambda_1\|\beta\|_1 \qquad (13)$$

Regarding IMAT, we first complete the matrix as in (12) and then we will proceed as in (5,7):

$$X^*(\lambda) = argmin\left(\left\|P_E(X - \hat{X})\right\|_2^2 + \lambda\|X\|_*\right) \quad (14)$$

followed by:

$$\beta_{k+1} = T_{k+1}(\beta_k + \lambda X^H(Y - X\beta_k)) \quad (15)$$

$$T_{k+1} = c \times \overline{\beta_k} \quad (16)$$

We will also provide how the methods behave if the data is precompleted. We use both high-rank and low rank data simulations in the results section. We divide the main data to the training and the test parts and apply the algorithms on the training data to learn the optimal parameters and the test data is used to find the RMSE estimation. In this paper, we assume that the size of training data is 0.8 of the size of main data ($\frac{4}{5}m$ rows) and size of the test data is 0.2 of the main data ($\frac{1}{5}m$ rows).

## IV. RESULTS AND SIMULATIONS

In this section, we provide our results and simulations. First, we explain the diverse types of generated data we employed. First, we generate the random data matrix by the idea of SVD. We form random orthogonal matrices $U, V$. Then, we generate random Gaussian singular values (based on the rank of the matrix) and finally form the matrix. We generate our parameters vector again in Gaussian format. Finally, we multiply these two and add noise with small variance to the entries. Next, we put a Bernoulli mask on the data matrix to induce missing samples into the structure of the problem. We have varied the size of our data matrix, the rank of the desired matrix, and also the level of missingness to have a comprehensive scrutiny over the Gaussian data. In Fig. 1, we clearly observe that the IMAT is performing stronger than the other methods. It is a low rank gaussian matrix with dimension 100*100 and the parameters are sparse with 8 non-zero entries. The horizontal axis shows the parameters of thethree implementations. It is worth noting that although the parameter for Lasso varies logarithmically, the parameters for IMAT and IHT are linearly swept, we include both curves on the same plot for the sake of comparison.

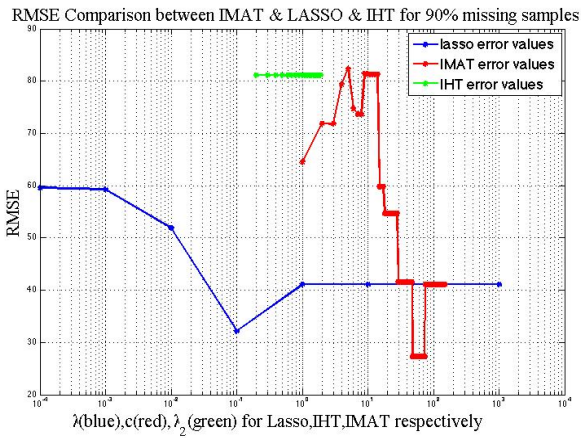

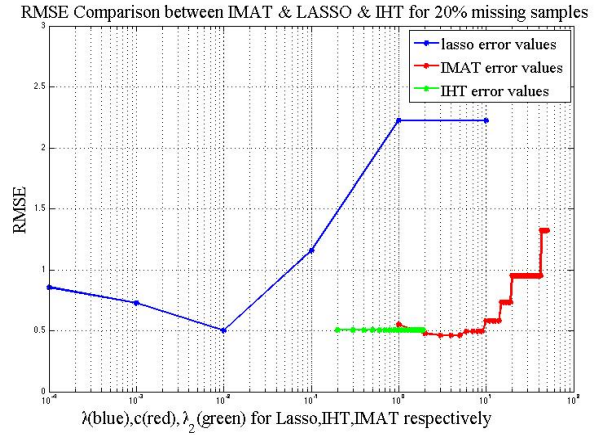

Fig. 1. Data is 100*100 with rank 50. the signal is sparse with 8 nonzero elements and 50 percents of the data is missing. The RMSE errors for prediction on test set after many trial are shown vs. the parameters of the three methods.

Fig. 2. Data is 1000*100 with rank 50. the signal is sparse with 8 nonzero elements and 20 percents of the data is missing. The RMSE errors for prediction on test set after many trial is shown vs. the parameters of the three methods. IMAT minimun RMSE= 0.4318, Lasso minimum RMSE=0.5243, IHT minimum RMSE= 0.5285.

We observe that the performance of all methods improve in terms of RMSE when we work with bigger size of data.

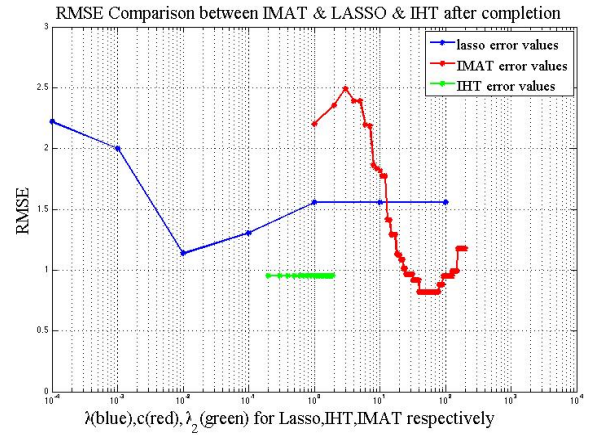

Fig. 3. Data is 1000*500 with rank 50. the signal is sparse with 8 nonzero elements and 50 perents of the data is missing. The RMSE errors for prediction on test set after many trial is shown vs. the parameters of the three methods. IMAT minimun RMSE= 0.8225, Lasso minimum RMSE= 1.1381, IHT minimum RMSE= 0.9547.

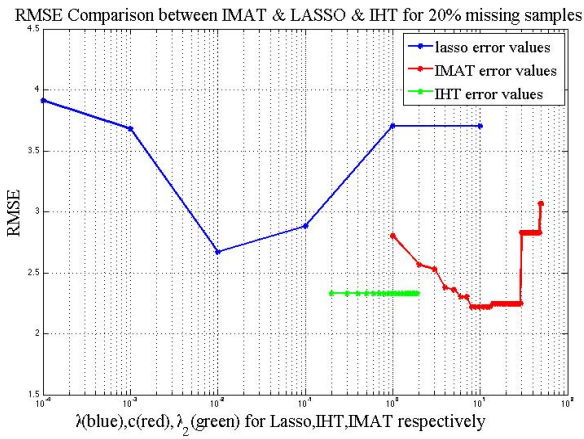

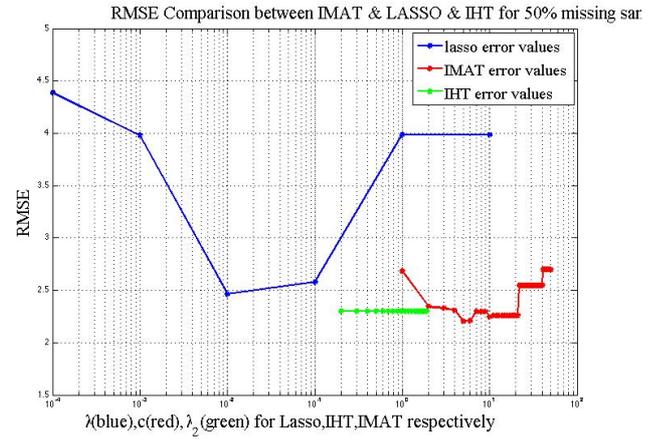

Fig. 4. Data is 1000*100 with rank 100. the signal is sparse with 8 nonzero elements and 50 percents of the data is missing. The RMSE errors for prediction on test set after many trial is shown vs. the parameters of the three methods. IMAT minimun RMSE= 2.1478, Lasso minimum RMSE=2.3135, IHT minimum RMSE= 2.2582.

Fig. 6. Data is 1000*100 with rank 20. the signal is sparse with 8 nonzero elements and 50 percents of the data is missing. The RMSE errors for prediction on test set after many trial is shown vs. the parameters of the three methods. IMAT minimun RMSE= 2.1102, Lasso minimum RMSE= 2.7147, IHT minimum RMSE= 2.2727.

Now we provide the results after matrix completion :
We observe that after completion the performance of Lasso is as optimal as IMAT. The issue with matrix completion is the time complexity of the algorithms to be implemented. If the purpose is to ignore the viable completion method and using the raw available data the power of data extraction for IMAT outperforms the LASSO. In order to provide a comprehensive comparison, we also plot the training runtime for the three approaches vs. the data size in Fig. 10. It is worth mentioning that the runtimes are obtained on a 2.7 GHz Intel Core i7 processor. It could be concluded form TABLE 1 that the proposed IMAT algorithm not only improves the reconstruction accuracy, but also it is more efficient in terms of runtime. Note that in the training step the parameter $\lambda$ is optimized logarithmically with exponential step size 10 in the interval [0.0001,100]. The IMAT parameter $c$ is optimized in a linear fashtion with step size 1. The general result is that in for the same number of parameters to learn IMAT requires less time to learn as well as improvinf the minimum RMSE achieved.

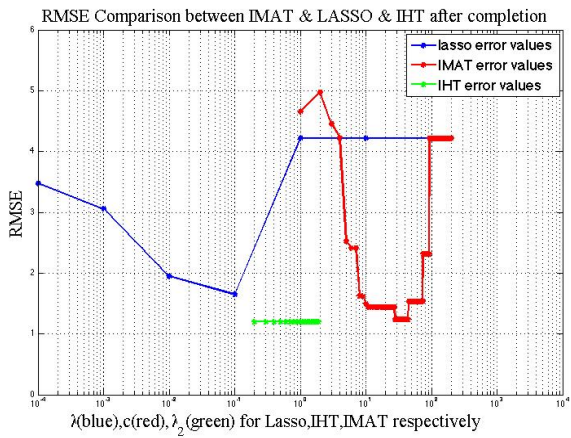

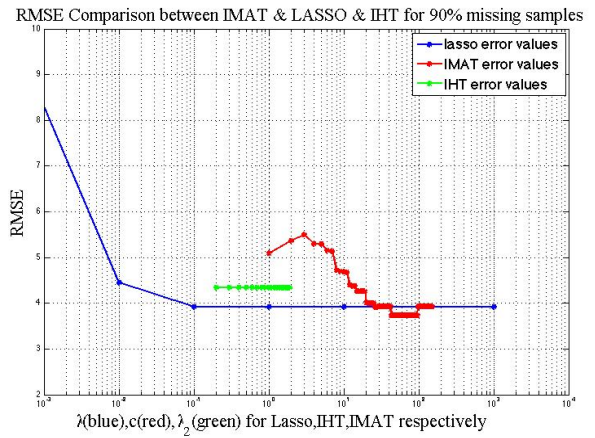

Fig. 5. Data is 500*200 with rank 50. the signal is sparse with 8 nonzero elements and 20 percents of the data is missing. The RMSE errors for prediction on test set after many trial is shown vs. the parameters of the three methods. IMAT minimun RMSE= 2.1478, Lasso minimum RMSE=2.3135, IHT minimum RMSE= 2.2582.

Fig. 7. Data is 1000*100 with rank 20. the signal is sparse with 8 nonzero elements and 90 percents of the data is missing. The RMSE errors for prediction on test set after many trial is shown vs. the parameters of the three methods.

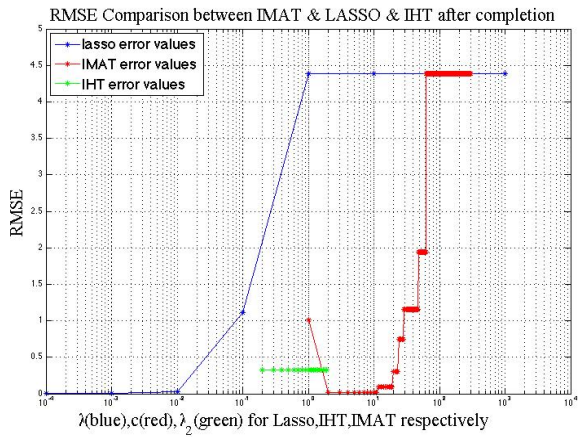

Fig. 8. Data is 500*200 with rank 50. the signal is sparse with 8 nonzero elements and 20 percents of the data is missing. Matrix completion is carried out. The RMSE errors for prediction on test set after many trials is shown vs. the parameters of the three methods.

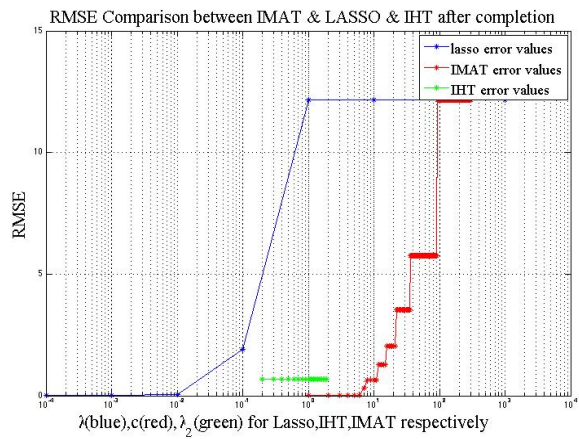

Data is 500*100 with rank 20. the signal is sparse with 8 nonzero elements and 90 percents of the data is missing. Matrix completion is carried out. The RMSE errors for prediction on test set after many trials is shown vs. the parameters of the three methods.

TABLE 1
Comparison between the runtimes achieved by LASSO and IMAT.

| Method / Data size | IMAT | LASSO |
|---|---|---|
| m= 100,n=100 | 0.0162 | 0.0307 |
| m= 200,n=100 | 0.0192 | 0.0351 |
| m= 500,n=100 | 0.0266 | 0.0227 |
| m= 1000,n=100 | 0.0300 | 0.0328 |
| m=2000,n=100 | 0.0304 | 0.0335 |
| m=1000,n=500 | 0.0584 | 0.1294 |

ACKNOWLEDGMENT *(Heading 5)*

CONCLUSION

We have found out that the IMAT has a better performance in recovering sparse signals in linear models than LASSO and IHT. We have observed in our diverse simulations which included various types of data matrices in terms of rank and missing samples that the RMSE of error for test set is less for IMAT in comparsion to other two methods and the gap of difference between the RMSE for IMAT and LASSO increases when the data is of low rank and smaller size. It also performs better than IHT which is dependent on knowing the sparsity of the desired signal. We have also noticed that the dominance of lasso is more observable when the size of data increases. It is also more efficient than lasso in terms of runtime and time complexity. We have tried random gerenrated data and found that IMAT works better for each scenario. We have found that the performance of the method in the presence of matrix completion is approximately similar to each other.